%% file: ms.tex
\documentclass[runningheads]{llncs}
\usepackage{graphicx}
\usepackage[utf8]{inputenc}
\usepackage{ dsfont }
\usepackage{libertine}
\usepackage{xcolor}
\usepackage{cite}
\usepackage{float}

\begin{document}
\title{Learning by Design: Structuring and Documenting the Human Choices in Machine Learning Development}
\titlerunning{Learning by Design}
%
%
\author{Simon Aagaard Enni\orcidID{0000-0002-1544-4371} \and
Ira Assent\orcidID{0000-0002-1091-9948} }
\authorrunning{S. A. Enni and I. Assent}
%
\institute{Aarhus University, Aarhus, Denmark}

\maketitle              
\input{Abstract}
\input{Introduction}

\input{Related_work}
\input{ML_problem_formulation}

\input{Design_choices}
\input{Conclusion}

\bibliographystyle{splncs04}
\bibliography{thesis}

\end{document}

%% file: Abstract.tex
\begin{abstract}
	
The influence of machine learning (ML) is quickly spreading, and a number of recent technological innovations have applied ML as a central technology. However, ML development still requires a substantial amount of human expertise to be successful. The deliberation and expert judgment applied during ML development cannot be revisited or scrutinized if not properly documented, and this hinders the further adoption of ML technologies--especially in safety critical situations.

In this paper, we present a method consisting of eight design questions, that outline the deliberation and normative choices going into creating a ML model. Our method affords several benefits, such as supporting critical assessment through methodological transparency, aiding in model debugging, and anchoring model explanations by committing to a \textit{pre hoc} expectation of the model's behavior. We believe that our method can help ML practitioners structure and justify their choices and assumptions when developing ML models, and that it can help bridge a gap between those inside and outside the ML field in understanding how and why ML models are designed and developed the way they are.

\keywords{Machine learning \and Documentation \and Development \and Methodology}
\end{abstract}

%% file: Introduction.tex
\section{Introduction}

Machine learning (ML) has led to some of the most influential innovations in contemporary computing even leading some to categorize ML as the beginning of a new data-driven industrial and scientific revolution \cite{Kit14}. With so much at stake, critical discussions about the proper role of ML in society and about the safety and ethics of ML technologies emerge \cite{Dan16, Hil16a, Mitetal16, One17, Zar12}. In ML, a machine \textit{learner} is developed using carefully chosen biases and data to search for a ML model, that performs some desired function, such as classification, rare-event detection, or clustering given previously unseen input data. While a ML model functions mostly automatically once learned, the process of creating an effective machine learner requires a lot of human labor and deliberation \cite{Dom12}. The main benefit of recognizing this indirect human influence is in making explicit the deliberation and design going into creating ML models, which allows us to systematize ML development, documentation, critical investigations, maintenance and revision. Recent critical discussions have revealed a serious need for such systematization, as critics liken the current state of ML to alchemy, based on "a substantial amount of 'black art'" and "folklore and magic spells" \cite{CamCra20}. The discourse surrounding ML as somehow magical and unknowable, especially coming from insiders in the field, is a risk to the professional and scientific integrity of the ML field and therefore should be taken seriously to avoid myth-making, unsubstantiated hype, and the inevitable resulting disappointment \cite{MosSch18}.
In the interest of the scientific and professional integrity of the ML field, we present a design method that illustrates the different kinds of design choices made by the developers of machine learners and associated design questions that guide these choices. In this way, we argue that ML models are not just learned, but also \textit{designed}, albeit indirectly.

Chief among the benefits afforded by our method is that it makes the process of creating machine learners more transparent, explicit, and interpretable, supporting critical assessment by noting design choices and their rationales in humanly readable form. It furthermore creates a documentation of the intended behavior of the ML model that serves as a \textit{pre hoc} explanation and justification, similar to the preregistration of an experimental setup \cite{Hil19}. This description can be used to anchor explanations of model behavior later on in the process and enables separating errors in implementation from errors in the design of the model, thereby aiding with debugging. At this level of abstraction, ML models can be approached by domain experts as well as ML specialists, allowing for tighter collaboration and discussion. 

As new legislation and research moves away from a reactive 'informed consent'-based paradigm towards protection by design, as outlined in article 25 of the GDPR,\footnote{Article 25 can be read at \url{https://gdpr-info.eu/art-25-gdpr/} - last accessed on 02-11-2020.} our method can furthermore be used to structure and document such protective design. Privacy by design \cite{Lan01}, fairness by design \cite{BarSel16, Dwoetal12}, and accountability by design \cite{Kroetal17, BerPre17} are all examples of this movement which could benefit from our method. Such approaches attempt to ensure that models are designed with values in mind that are aligned with those of society more broadly, illustrating the necessity of a structured and explicit design strategy for ML models.

%% file: Related_work.tex
\section{Related work}

There have been attempts to organize the workflow of ML development and the related discipline of data mining, that has resulted in process models \cite{Rei97, She00, FayPiaSmy96, Ameetal19, Hiletal16}. Our contribution in this paper differs from those by focusing not on how to organize a team of ML developers, or on outlining the different phases of development for ML systems, but instead on documenting, justifying, and deliberating on the model- and system-specific design choices made in ML development. Our model distinguishes different technical design choices and their associated design questions; regardless of workflow and makes different delineations as a result. 

The work of Yoram Reich in applying ML to civil engineering problems deserves a special mention \cite{Rei97}. While his method also focuses on ML modeling, it has a different application and focus. First off, Reich applies ML model to solve civil engineering problems, and thereby his analysis focuses on matching ML methods to engineering problems. Our approach is instead created as an aid in documenting and deliberating on the design of and assumptions made when creating general ML models.

The field of software testing has recently seen an upsurge of research in ML testing with the purpose of detecting ML "bugs", defined as "the differences between existing and required behaviors of an ML system" \cite{Zhaetal20}. In a related vein, explainable AI (xAI) has also seen a resurgence in interest, among other things as a common-sense driven way of evaluating the reasonableness of ML models' behavior through manual inspection \cite{Lip18, DosKim18}. Our contributions are complementary to both these fields, as the documentation provided through our method can aid both in identifying the required behaviors of an ML system for the ML testing field and in producing \textit{pre hoc} explanations for systems that can contextualize later \textit{post hoc} explanations from xAI.

Mitchell et al. recently presented a principled documentation technique for ML models in the form of Model Cards \cite{Mitetal19}. Models Cards are focused on documenting facts about finished models, in particular performance and fairness properties, whereas our model is made to document and guide the \textit{design} of the model as it is being developed. However, the documentation derived from our method can be added to a Model Card, making the methods complementary.

%% file: ML_problem_formulation.tex
\section{Approaching a Problem with ML: Pitfalls and Promises}
\label{sec:problem}

Perhaps the greatest strength of ML is that it enables programming-by-example, allowing for models to be produced for tasks where algorithmic solutions either do not exist or are insufficient for practical use. Instead, a description of the task is given along with examples demonstrating the behavior that the model should learn to mimic for the given task. T. Mitchell thus presented this canonical definition of ML:
\begin{quotation}
	A computer program is said to \textbf{learn} from experience $E$ with respect to some class of tasks $T$ and performance measure $P$, if its performance at tasks in $T$, as measured by $P$, improves with experience $E$. \cite[p. 2]{Mit97}
\end{quotation}

For learning to be possible, $T$, $E$, and $P$ must be carefully designed and quantified such that the abstract pedagogical issue of 'learning' is reduced to the concrete problem of iteratively and quantitatively improving a measure, $P$, with more experience, $E$, i.e., an optimization problem. It is here we see the necessity of the design aspect of ML; it is through design that the specifics of the task are described and made "learnable."

In the following we make an important distinction between a machine \textit{learner} and a ML \textit{model}. A machine learner is a program implemented and designed by a human programmer that produces a ML model through a process of learning from training data. It is the design of the machine learner that we are concerned with in this paper, as this design indirectly determines the functionality and behavior of the ML model.

However, solving problems by learning solutions introduces pitfalls as the task that the machine learner is meant to learn a model for is typically \textit{under-specified}. This is a natural result of trying to learn a general model from examples and has to be handled by carefully chosen assumptions. Poor choices here might lead to the resulting ML model being \textit{misaligned} with the intention of the developers in a number of ways; such as by displaying a high rate of error, making socially discriminatory decisions, or relying on unstable or spurious correlations. However, as mentioned previously, assumptions in ML are often poorly documented, instead deferring to professional hunches, best practices, and "black art" when making the decisions. In Section \ref{sec:design}, we will present a model outlining the different design choices made when designing a machine learner, which can help both those inside and outside the ML field understand and communicate these choices, shedding light on the "black art" of ML.

%% file: Design_choices.tex
\section{The indirect design of machine learning models}
\label{sec:design}

\begin{figure}[h]
	\centering
	 \makebox[\textwidth][c]{\includegraphics[width=1.1\textwidth]{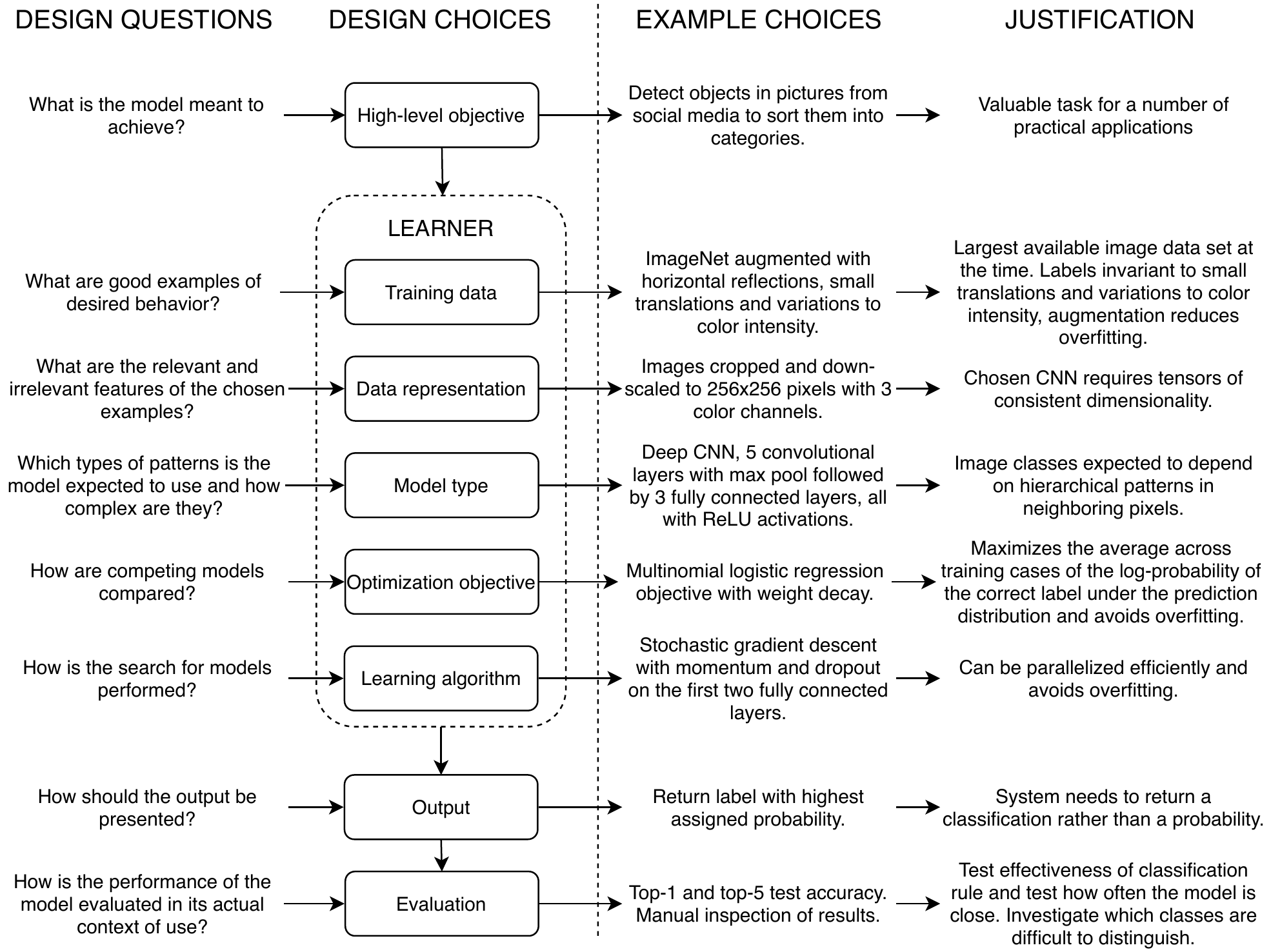}}
	\caption{An illustration of the different design choices made in the creation of a ML model and the design questions they are responses to. The first two columns on the left outline the design questions and the two columns on the right show the particular choices in our running example with corresponding justifications.}
	\label{fig:design_choices}
\end{figure}

In this section, we present a method outlining eight different categories of design choices that are made in the development of a machine learner, five of which spell out the technical make-up of the learner itself. Each of these represent a response to a particular question about the desired functionality of the ML model and can be seen in Figure \ref{fig:design_choices}. The remainder of this section will cover each design choice in turn. In order to make our points more concrete, we will introduce a running example of a fictitious ML system designed for object detection where a design choice is made and justified in each section. We choose to model this example on the well-documented AlexNet \cite{KriSutHin12}, but we stress that our example is fictitious, and that the example motivations and deliberations for the individual design choices are our own for purely illustrative purposes. The two right-most columns in Figure \ref{fig:design_choices} display an overview of the choices and their justifications in this example.

\subsection{High-level objective}
\label{sec:high_level_objective}

\begin{quote}
	\textit{What is the model meant to achieve?}
\end{quote}

The high-level objective is the task that the ML model is meant to perform once completed, and it motivates the rest of the design choices. This objective is not necessarily meant to be a technical description, but rather a declaration of the intent of the developers in designing the ML model. It can be abstract, such as detecting objects in images or it could be very specific, such as finding anomalies in the water consumption data of a specific supermarket chain to detect faults and reduce water consumption \cite{Veretal18}. The high-level objective has to be specific enough to support decision making in other sections towards achieving this goal. An objective such as "detecting objects in images" may be too vague in this regard, as it is not even clear which kind of objects should be detected in which kind of images, let alone for which purpose these objects should be detected. The anomaly example above \cite{Veretal18} is more specific and lends itself well to documenting a ML design process. Since the kind of data and its source is known and the purpose of the finished system is already declared, a lot of the following design choices can be clearly motivated. While it is not known in advance what constitutes an anomaly in the data, it is known what these anomalies would be used for, so the search can be focused on those anomalies that, e.g., can be related to faults and excessive water consumption.

Ideally the objective should specify (1) what kinds of data the model is expected to work with, (2) what task the model is expected to perform, (3) why the model is made for that task, and (4) which context the model is expected to be deployed in. While every ML project is motivated by some high-level objective, such objectives often change and are iteratively improved as the project progresses and the feasibility of the objective is tested \cite{PasSen20}. 

\textit{In our running example, we choose to design a ML model for detecting common objects in pictures in order to classify and categorize pictures from social media based on their contents. The model is thus expected to work with digital photographs of varying resolution, and we would expect that the model should be deployed on personal computers to help regular users of social media. }

\subsection{Training data}
\label{sec:training_data}

\begin{quote}
	\textit{What are good examples of desired behavior?}
\end{quote}

Ideally, the training data will supply sufficiently diverse examples to eliminate any other models than the ones that work as intended. Furthermore, in order to successfully generalize to new situations, a lot of data are often needed, as the confidence and accuracy of model predictions tend to improve with more data \cite{Mit97,HasTibFri09}. High diversity eliminates models built on spurious patterns and large amounts of data allow for better statistical bounds on the expected error of the resulting model. Training data are chosen to exemplify desired behavior in the learned model, and the resulting model may become misaligned if the data are not sufficiently representative of this behavior.

There has long been awareness of the dangers of poor data \cite{Kimetal03}, and the adage "garbage in, garbage out" is often used to describe the futility of doing good analysis with bad data. The data are not just meant to be free from errors, but also to clearly illustrate the task that the ML model is meant to learn. Explicitly asking "what are good examples of desired behavior" as part of the design process encourages reflection and enables critical scrutiny and stakeholder inquiries. The chosen training data are contextualized as deliberately selected information with an explicit purpose.  

\textit{In our running example, we choose to use the ImageNet data set \cite{Rusetal15}, as it constitutes one of the largest collections of pictures tagged with the objects depicted in them. ImageNet contains millions of pictures scraped from different websites classified into hundreds of categories, and we expect these pictures to be representative of what the model would encounter when deployed in practice. We furthermore augment the data by introducing horizontal reflections, small translations, and variations to color intensity, as we expect the class of each picture to be invariant to such transformations, and such augmentations are expected to reduce overfitting \cite{KriSutHin12}. }

\subsection{Data representation}

\begin{quote}
	\textit{What are the relevant and irrelevant features of the chosen examples?}
\end{quote}

Once the training data are chosen they have to be represented in a way that facilitates learning. This representation serves the purpose of highlighting what information is relevant to the task and removing distractions. Often, the data used in ML are represented as \textit{feature vectors} via. \textit{feature engineering} \cite{Dom12, LecBenHin15} where a series of quantified vector representations are tested through a process of trial and error. However, the data may be represented in any number of ways that the developer believes illustrates their important aspects well, such as tensors for image and video data which preserve the locality of pixel data, or graphs for network data, which preserve and highlight the structure of the analyzed network.

A well-designed data representation excludes irrelevant information while retaining as much relevant information as possible. However, what counts as relevant and irrelevant information is not defined \textit{a priori}. Recent advances in representation learning have partially automated this process, when the relevant information can be described with more abstract biases \cite{LecBenHin15}. However, it is important to carefully consider which features would facilitate the learning process and which would hinder it. If important information is removed by mistake, or if too much irrelevant information drowns out the relevant bits, the model may learn spurious correlations between irrelevant features. Furthermore, choices of model type and data set naturally constrain what kinds of data representations are feasible for the particular task, as the chosen representation has to fit them both.

\textit{In our running example, choice of data and model type indeed impacts representation. We choose to represent the data as tensors with 256x256 pixels and 3 color channels owing to two primary considerations. First, we use a convolutional neural network (CNN) for classification, and such models require input of a fixed dimensionality. Second, the tensor representation preserves the relative location of pixels, which we expect to be beneficial to detecting objects that extend over adjacent pixels.}

\subsection{Model type and architecture}
\label{sec:model_type}

\begin{quote}
	\textit{Which types of patterns is the model expected to use and how complex are they?}
\end{quote}

For any data set there are many potential models consistent with the data, and choosing any one model over the rest requires specifying an \textit{a priori} preference in the form of an \textit{inductive bias} \cite{Mit97}. With a bad inductive bias, the chosen model might generalize in ways that are not aligned with the interests of the developers even if they are consistent with the training data. Such misalignment might show itself merely as bad performance, but it may also be more subtle and problematic such as relying on unstable or spurious correlations \cite{CalLon17} or making socially discriminatory generalizations \cite{BarSel16}. The first part of the inductive bias is given by the model type and architecture, and the second part is discussed in Section \ref{sec:opt_obj}. 

Choosing a good model type comes down to reasoning about the nature of the relation between input and output in the chosen task. While any ML project implicitly assumes the existence of some statistical regularity governing the relation between input and output, choosing a particular model type biases the learner towards learning specific kinds of patterns such as those that can be represented by a weighted sum of the feature values in the case of a linear model. Assumptions of linearity may not hold for many problems and often more advanced models with subtler biases are used to represent complex dependencies between input and output. Increased complexity comes at a cost, however, both with respect to the amount of data required for successful learning and the number of computations needed at each step in the learning process \cite{Mit97, HasTibFri09}. Thus, the computational budget available might become a deciding factor in the choice of model type and architecture, and wherever this is the case, it is important to document such decisions, so that they may be reevaluated if the availability of computational resources change.

A poor choice of model type and architecture may result in the model being misaligned with the developers' intentions if important patterns in the input-output relationship cannot be captured by the model. If sequential order of the data is paramount, such as in speech recognition, but a model type with an assumption of no such sequences in the data is used, the model will most likely fail to capture anything but weak and spurious patterns. This shows the importance of careful design, as a good choice of model type and architecture depends strongly on what is known about the relation between input and output and which patterns to expect in the data. Documenting the reasoning behind the model design by answering which patterns are expected in the data and which structure the model is expected to exploit allows later studies to critically scrutinize this fundamental aspect. For example, CNNs trained on ImageNet were investigated on account of the assumption that they learn hierarchical representations of increasingly complex shapes. Instead, experiments exposed a texture bias in the models that undermined this assumption \cite{Geietal19}.

\textit{In our example, we choose a CNN with 5 convolution layers designed to detect a hierarchical representation of pixel patterns as we expect that many features central to object recognition such as compact shapes and textures are defined by correlations in neighboring pixels. These layers are followed by 3 fully connected layers to learn a classification function from the features detected by the convolutional layers. ReLU activations are used throughout, as these have been shown, e.g., to avoid exploding gradients \cite{KriSutHin12, GloBorBen11}}.

\subsection{Optimization objective}
\label{sec:opt_obj}

\begin{quote}
	\textit{How are competing models compared?}
\end{quote}

While the chosen model type and architecture determine which general patterns are expected, constituting the first part of the inductive bias, the optimization objective determines how different parameterizations of the same model are compared, constituting the second part of the inductive bias. Choosing the optimization objective translates learning into a problem, where each parameterization of the model maps to an associated value. In other words, it determines how different instantiations of the same model can be ranked and thereby it determines what is considered a successful model. In order for the translation from abstract task to concrete objective to be successful, the optimization objective should account for the different kinds of errors the model can make and their relative values as well as any constraints the parameters of the model are subject to. However, as in the choice of model type, the optimization objective also has a computational aspect to it. 

It is worth noting that the optimization objective can account for more things than just predictive power, e.g., classifiers can be optimized for increased fairness \cite{EnnAss18}, or they can be optimized to shift attention to patterns considered more sensible \cite{RosHugDos17}. If important goals or constraints are not accounted for in the optimization objective, the learner might output a model that is inconsistent with those. 

\textit{In the running example we choose a multinomial regression objective with weight decay regularization. With this choice the model optimizes its chances of making the correct predictions across the training set, and the regularization reduces overfitting due to the large number of parameters in the model \cite{HasTibFri09}.} 

\subsection{Learning algorithm}

\begin{quote}
	\textit{How is the search for models performed?}
\end{quote}

Where the optimization objective defines criteria to evaluate the relative value of model parameterizations, the learning algorithm decides how the search for the optimal model is performed. It does not, however, imply any behavior for the model in its own right. The learning algorithm is evaluated on its efficiency and effectiveness--the parameters of the optimal model do not change based on the method used to arrive at them. Two different learning algorithms might still arrive at parameterizations with different performance and at a different computational cost, though, as there are often no guarantee that the learning algorithm arrives at the single global optimum. 

However, since the efficiency of the learning algorithm is an important factor in deciding the computational cost of training a ML model, it is not uncommon to revisit other design choices to make them compatible with a powerful learning algorithm, even if they represent the task less accurately as a result. This is the case, e.g., for gradient descent where the loss-function, and thereby also the model's prediction function, must be differentiable, which also constrains the input data to be a vector, matrix, or tensor of fixed dimensionality. An example of compromise could be choosing a convex optimization objective that less accurately represents the targeted problem, but which guarantees that gradient descent converges towards the global optimum. By explicitly noting such assumptions and considerations going into the choice of a particular learning algorithm it can easily be revisited and adjusted if necessary.

\textit{In the running example we choose to train the network using stochastic gradient descent with momentum, since this method has previously shown good performance on similar CNN models and other deep neural networks \cite{Sch15}. We add dropout \cite{Hinetal12} to the first two fully connected layers to further reduce the risk of overfitting by forcing neurons to learn to detect features that do not require the presence of particular other neurons \cite{KriSutHin12}.}

\subsection{Output}

\begin{quote}
	\textit{How should the output be presented?}
\end{quote}

In order to perform its task, the model's output often has to be changed to some useful format. The model type and/or optimization objective often requires a continuous output from the model to facilitate optimization, and this output can be interpreted, e.g., as the probability of seeing the positive class, in the case of a logistic regression \cite{HasTibFri09}. However, what is needed from a model is often classification rather than a distribution of probability over classes, and the output must be post-processed to achieve this functionality. In other cases the numeric output might have to be changed to, e.g., a risk scale of low, medium and high \cite{ProPubCOMPAS16Analysis}. 

Thus a choice has to be made as to how the output is presented to the user and while it does not affect the behavior of the model directly, it might have important implications for its functionality. In particular it affects how the system can be tested, as the form of the output might decide which outputs are errors and which are not.

\textit{In our running example we make a simple modification by outputting the class with the highest predicted probability as the predicted class. We make this change because the system is meant to be used for classifying images, and this requires returning a single predicted class for each input.}

\subsection{Performance evaluation}

\begin{quote}
	\textit{How is the performance of the model evaluated in its actual context of use?}
\end{quote}

The final choice in creating a ML model is the evaluation scheme gauging the performance of the trained model. Evaluating a ML model on held-back validation data often informs and guides the design of its next iteration, and thereby directly affects the design of the model \cite{HasTibFri09, Lan11}. However, the importance of a proper \textit{experimental design} is often underappreciated, especially when experimental designs are inherited from previous projects \cite{Fla19}.

Two important aspects must be be given particular attention: the quality of the test data and the choice of evaluation metric. Ideally, the testing data would be an unbiased and representative sample of the data the model is expected to encounter in practice to get as good of an estimate of performance as possible \cite{Mit97}. Thus, getting a testing data set of high quality requires intimate knowledge of the problem that the model is made to address to account for label skews, biases, concept drift and more. To ensure that all such potential problems are accounted for it may be necessary to measure test data oneself, if budget and practical considerations allow for it. 

The second issue can become similarly subtle, as choosing a good evaluation metric depends on the intended use-case of the learned model \cite{Fla19}. Knowing how and for what the model is going to be used is necessary to choose a measure that makes sense for this purpose. The performance of a ML model might also have several qualitatively different facets that cannot be meaningfully aggregated into a single measure of performance such as recall and precision. Another example of this problem is when a model is required to be "fair", which requires choosing and justifying a particular measure of fairness from a vast population of fairness measures \cite{CorGoe18}.

\textit{In the running example, we measure the top-1 accuracy, to test the efficiency of the classification rule chosen in the output presentation, and the top-5 accuracy to test how often the model is close to being right, both measured on held-back testing data. We will also manually inspect a random sample of the top-5 results in order to see which specific classes the model has trouble distinguishing. These measures relate to the high-level objective as we wish to know which picture classes will likely contain the most errors and if the classification could improve if we return more than one label per picture.}  

%% file: Conclusion.tex
\section{Conclusion and discussion}
\label{sec:conclusion}

We presented a method consisting of eight design questions, the answers to which serve to describe the desired behavior of the ML model learned by a machine learner. This method can shed light on the deliberation and the normative choices that are made when constructing machine learners. By elevating the design process to the level of abstraction concerning the properties of the input-output relation modeled by the ML model rather than the level of technical and mathematical specifics, this method might make ML more approachable to actors both inside and outside the ML field, thereby supporting a higher degree of collaboration and discussion. 

By noting the design choices and their rationales, the desired ML model is described in humanly readable terms and ML developers thereby produce a \textit{pre hoc} explanation of the expected behavior and justification of the model, similar to the preregistration of a scientific experimental setup \cite{Hil19}. This description can help making expectations concrete and tangible and can anchor later \textit{post hoc} explanations \cite{Guietal18b} of model behavior. Describing the desired model by iteratively refining answers to the eight design questions, generates a stated justification which can act as an anchor for discussions and maintenance. It thereby gives a lens through which to explain, debug, or contest the model, opening up a discussion not just of the effectiveness of these choices but also of their appropriateness.